\pdfoutput=1

\documentclass[11pt]{article}

\usepackage[preprint]{acl}

\usepackage{times}
\usepackage{latexsym}
\usepackage{listings}
\usepackage{amsmath}
\usepackage{enumitem}
\usepackage{booktabs}
\usepackage{tabularx}
\usepackage{graphicx}   
\usepackage{subcaption} 

\usepackage[T1]{fontenc}

\usepackage[utf8]{inputenc}

\usepackage{microtype}

\usepackage{inconsolata}

\usepackage{graphicx}
\usepackage{booktabs}
\usepackage{xcolor}
\usepackage[normalem]{ulem}
\useunder{\uline}{\ul}{}
\usepackage{tabularx}
\usepackage{amsmath}
\newcolumntype{Y}{>{\centering\arraybackslash}X}

\title{Modeling Transformers as complex networks to analyze learning dynamics}

\author{
\textbf{Elisabetta Rocchetti\textsuperscript{1}},
\\
\textsuperscript{1}Università degli Studi di Milano, Department of Computer Science, Via Celoria, 18 - 20133 Milan, Italy\\
\small{
\textbf{Correspondence:} elisabetta.rocchetti@unimi.it
}
}

\begin{document}
\maketitle

\begin{abstract}
The process by which Large Language Models (LLMs) acquire complex capabilities during training remains a key open question in mechanistic interpretability. This project investigates whether these learning dynamics can be characterized through the lens of Complex Network Theory (CNT). I introduce a novel methodology to represent a Transformer-based LLM as a directed, weighted graph where nodes are the model's computational components (attention heads and MLPs) and edges represent causal influence, measured via an intervention-based ablation technique. By tracking the evolution of this component-graph across 143 training checkpoints of the Pythia-14M model on a canonical induction task, I analyze a suite of graph-theoretic metrics. The results reveal that the network's structure evolves through distinct phases of exploration, consolidation, and refinement. Specifically, I identify the emergence of a stable hierarchy of \textit{information spreader} components and a dynamic set of \textit{information gatherer} components, whose roles reconfigure at key learning junctures. This work demonstrates that a component-level network perspective offers a powerful macroscopic lens for visualizing and understanding the self-organizing principles that drive the formation of functional circuits in LLMs.
\end{abstract}

\section{Introduction}
Large Language Models (LLMs) have demonstrated remarkable capabilities, yet their internal workings remain largely opaque. The field of mechanistic interpretability aims to reverse-engineer the algorithms these models learn, moving from correlational observations to a causal understanding of their computations. A key emergent phenomenon is in-context learning (ICL), where a model adapts to a new task at inference time based on examples in its prompt. Seminal work has shown that specific circuits, such as induction heads, are crucial for ICL and that they form during a distinct ``phase change'' early in training~\cite{olsson2022incontextlearninginductionheads}.

While the microscopic details of such circuits are beginning to be understood, a macroscopic perspective on the model's overall architectural dynamics during these phase changes is still missing. Complex Network Theory (CNT) offers a powerful framework for this, having been successfully applied to analyze the structure and performance of other neural networks~\cite{you2020graphstruc, chen2025biologically}. However, previous applications to Transformers have primarily focused on creating token-level graphs from attention patterns to understand how the model represents input data~\cite{zheng2024chinese}. A critical, unexplored dimension is how the internal \textit{computational components} of an LLM---the attention heads and MLPs---organize themselves into a functional network via the residual stream.

This project aims to fill this gap. My central hypothesis is that training is associated with distinct and measurable structural signature in a network of the model's components. To test this, I introduce a novel methodology to model a Transformer as a component-level graph where edges are defined not by parameter weights, but by the causal influence one component has on another's output. By tracking the evolution of this graph's topological properties across training, I provide the first analysis of how a model's internal communication architecture self-organizes to learn a specific task. This work demonstrates the feasibility of using CNT as a new tool for mechanistic interpretability, offering a complementary, macroscopic view on the formation of functional circuits.

The remainder of this project is as follows. Section~\ref{sec:related_work} provides context by reviewing prior work on modeling neural networks with CNT. Section~\ref{sec:preliminaries} establishes the necessary background on Transformer architecture, mechanistic interpretability, and in-context learning. My method for translating a Transformer into a component-level graph is detailed in Section~\ref{sec:methodology}, followed by the specific experimental protocol in Section~\ref{sec:exp-setup}. In Section~\ref{sec:results}, I present the core findings, analyzing the evolution of the graph's structure and correlating it with the model's learning dynamics. Finally, Section~\ref{sec:conclusion} concludes the project by summarizing the key insights and discussing the limitations and future directions of this work.

\section{Related work}
\label{sec:related_work}
The analysis of Artificial Neural Networks (ANNs) through the lens of CNT has emerged as a powerful paradigm for understanding the relationship between a model's structure and its predictive capabilities. This line of inquiry seeks to uncover principles of information flow and organization that govern learning dynamics, drawing parallels with biological neural networks (BNNs).

\subsection{Neural networks as graphs}
Initial foundational work established the viability of representing ANNs as graphs. \citet{you2020graphstruc} introduced the pioneering concept of a \textit{relational graph}, where nodes represent feature channels and edges model the message-passing between them. By analyzing metrics such as the clustering coefficient and average path length, they made two seminal discoveries: first, that a network's performance is a smooth function of these graph properties, with optimal models occupying a ``sweet spot'' of structural balance; and second, that these top-performing ANNs share striking structural similarities with real BNNs. Building on this, \citet{scabini2023structure} shifted the focus from global graph properties to local neuron-level characteristics. Analyzing fully connected networks, they demonstrated that measures of neuronal centrality strongly correlate with performance and introduced the ``Bag-of-Neurons'' concept to identify distinct topological roles that neurons adopt during training, noting that poor performance was linked to neuron saturation indicated by excessively high subgraph centrality.

\subsection{Networks' training through CNT}
While these studies focused on the static topology of trained networks, subsequent research began to investigate the dynamics of network formation during training. \citet{lamalfa2021characterizing} developed a framework to track learning dynamics in real-time using CNT metrics like link weights, node strength, and layer fluctuation, providing a window into how information flow evolves. This was further advanced in their later work~\citep{lamalfa2024dnnperspective}, which introduced crucial data-dependent metrics like neurons strength and neurons activation. This extension allowed the analysis to move beyond pure topology and account for how input data actively shapes the network's functional graph, revealing unique architectural signatures for CNNs, RNNs, and Autoencoders.

More recently, this graph-based methodology has been adapted to the transformer architecture. Applying these concepts to Vision Transformers (ViTs), \citet{chen2025biologically} proposed a dual-component graph consisting of an aggregation graph for token-level spatial interactions and an affine graph for channel-wise communication. They reaffirmed the existence of a ``sweet spot'' for ViTs and demonstrated that their graph structures are even more similar to advanced mammalian BNNs than those of CNNs.

\subsection{Previous works vs my contribution}
These recent transformer-focused studies primarily construct graphs where tokens are the fundamental nodes, aiming to understand how the model represents input data. A critical, unexplored dimension is how the internal computational components of a language model—the attention heads and MLPs—organize themselves into a functional network. How these components communicate via the residual stream and how this communication network evolves during the learning of specific capabilities, such as in-context learning, remains an open question.

This project fills this gap. I propose a novel approach to model LLMs as complex networks where the nodes are the architectural components themselves. Instead of using direct parameter weights as edges, I define connections based on a form of intermediate attribution, quantifying how components causally influence each other's outputs. By tracking the evolution of this component graph during training on an induction task, my work provides a first analysis which, in future works, can reveal how functional circuits, like induction heads, emerge from the perspective of network dynamics.

\section{Preliminaries}
\label{sec:preliminaries}
To provide the necessary background for my methodology, this section reviews the architecture of auto-regressive Transformer-based LLMs, and the specific phenomenon of in-context learning driven by induction heads.
\subsection{Transformer architecture}
We can express the auto-regressive language modeling process performed by an LLM as the probability of a token sequence $T = (t_1, t_2, \dots, t_K)$:
\begin{equation}
\label{eq:lm}
  P(T) = P(t_1, t_2, \dots, t_K) = \prod_{i = 1}^{K}p(t_i|t_{<i})
\end{equation}
where $t_i$ represents the token at position $i$, $t_{<i}$ denotes the sequence of preceding tokens $(t_1, \dots, t_{i-1})$, and $K$ is the total number of tokens in the generated sequence $T$.

In a Transformer-based LLM, $p(t_i|t_{<i})$ is derived from the final hidden state as computed during the $i$-th generation step. The model architecture consists of an embedding layer, $L$ stacked Transformer blocks, and a final output layer.
Let $\mathbf{h}^{(0)}$ be the input token embeddings (plus positional embeddings). The $l$-th Transformer block, denoted $b_l$, processes an input $\mathbf{h}^{(l-1)}$ (the output from the preceding block) to produce $\mathbf{h}^{(l)}$. Each block $b_l$ typically comprises a multi-head self-attention module ($A_l$, taking as input $\mathbf{h}^{(l-1)}$), a feed-forward network ($F_l$, taking as input $\mathbf{h}^{(l-1)}_{\text{\textit{post-Attn}}}$, which is the output of $A_l$), and layer normalization ($LN$, which takes as input $F_l$'s output). Each block computes $\mathbf{h}^{(l)}$ as 
\begin{align*}
  b_l(\mathbf{h}^{(l-1)}) = LN(F_l(\mathbf{h}^{(l-1)}_{\text{\textit{post-Attn}}}) + \mathbf{h}^{(l-1)}_{\text{\textit{post-Attn}}})\\
  \mathbf{h}^{(l-1)}_{\text{\textit{post-Attn}}} = LN(A_l(\mathbf{h}^{(l-1)}) + \mathbf{h}^{(l-1)})
\end{align*}
The final output layer then takes $\mathbf{h}^{(L)}$ and projects it back to the vocabulary space using the unembedding weight matrix $W_U$:
\begin{math}
  \mathbf{h}^{(L)}W_U
\end{math}.

\subsection{Induction heads}
A key capability of modern LLMs is In-Context Learning (ICL), the ability to perform new tasks or adapt behavior at inference time based solely on examples provided in the input prompt, without any updates to the model's parameters. A primary mechanistic explanation for this phenomenon is the emergence of \textbf{induction heads}, specialized circuits that enable the model to perform pattern completion from the context~\cite{olsson2022incontextlearninginductionheads}. An induction head is not a single component but rather a circuit formed by two interacting attention heads (e.g., $A_l$ and $A_{l'}$ with $l<l'$). Mechanically, this circuit operates by identifying a token in the current sequence (e.g., $t_1$), searching for its previous occurrences, finding the token that followed it in the past (e.g., $t_2$), and then strongly predicting $t_2$ to follow $t_1$ again. This allows the model to continue sequences it has seen before within the same context, forming the basis of its ICL ability.

The formation of these critical circuits is tightly linked to a distinct \textbf{phase change} that occurs early in the training process, observable as a characteristic ``bump'' in the training loss curve. This phase change marks a pivotal moment in the model's development: it is precisely during this period that the LLM simultaneously forms its induction heads and acquires the vast majority of its in-context learning capabilities. Therefore, this phase change can be understood as the transition where the LLM develops the mechanistic circuitry required for more complex, context-dependent skills, moving beyond simple token co-occurrence statistics.

\section{Methodology}
\label{sec:methodology}
To investigate my hypothesis that an LLM's learning dynamics correspond to an evolving complex network, this methodology first translates the Transformer into a component-level graph at each training checkpoint. I then apply complex network metrics to analyze the training process, relating the graph’s structural evolution to the model’s performance.

\subsection{Graph representation of a Transformer}
I construct a directed, weighted graph $G=(V, E)$ where the set of vertices $V$ represents the core computational units of the Transformer: its attention heads ($A_l$) and MLP blocks ($F_l$) for all layers $l \in \{1, \dots, L\}$. This component-level view differs fundamentally from prior literature that uses tokens or individual neurons as nodes. My rationale is that these components---attention heads and MLPs---are the fundamental units that learn to process and route information through the model. The edges of this graph are designed to capture the non-trivial communication between these units that occurs via the residual stream.

An edge is drawn from an earlier component $C_i$ to a later component $C_j$ (where the layer of $i$ is less than the layer of $j$) if $C_i$ has a significant causal influence on the output of $C_j$. I quantify this influence using an intervention-based approach. For a given input sequence, I perform two forward passes: (1st pass) a \textbf{clean} run, which is a standard forward pass, where I record the output of component $C_j$, denoted as $O_j^{\text{clean}}$; (2nd pass) an \textbf{ablated} run, which is a forward pass where the output of component $C_i$ is zero-ablated before it is added to the residual stream. I then record the new, perturbed output of $C_j$, denoted as $O_j^{\text{ablated}}$.

The influence is measured by the change in $C_j$'s output, quantified by the cosine similarity $S$ between the clean and ablated outputs. A directed edge from $C_i$ to $C_j$ exists if this similarity drops below a predefined threshold $\tau$:
\begin{equation}
    S(O_j^{\text{clean}}, O_j^{\text{ablated}}) < \tau
\end{equation}

The weight of the edge $(i, j)$, denoted $w_{ij}$, is defined to be proportional to the magnitude of the impact. A stronger impact corresponds to lower cosine similarity. Therefore, the weight is set to:
\begin{equation}
    w_{ij} = 1 - S(O_j^{\text{clean}}, O_j^{\text{ablated}})
\end{equation}
This formulation ensures that stronger causal connections are represented by higher-weighted edges in the graph. The resulting graph is constructed for the model at each saved checkpoint, allowing to study its evolution throughout the training process.

\subsection{Complex network metrics}
I analyze the constructed graphs using a set of standard complex network metrics to characterize their structure and its evolution.

\paragraph{Network size and density}
Basic topological properties provide a high-level view of the network's connectivity. I measure the number of \textbf{nodes} $|V|$, number of \textbf{edges} $|E|$, and the network \textbf{density} $D = \frac{|E|}{|V|(|V|-1)}$. Density, in particular, indicates the overall level of component interaction. A high density suggests that information from most components is broadly utilized, whereas a low density implies the formation of more specialized communication pathways. The distribution of edge weights further clarifies this: a long-tailed distribution would suggest a network organized around a few strong communication channels (like highways) and many weaker ones (local streets).

\paragraph{Weighted in-degree and out-degree}
I measure the weighted in-degree and out-degree for each node to identify key components in the network. A high \textbf{out-degree} signifies a component that acts as an information ``spreader,'' whose output is crucial for many subsequent components. Conversely, a high \textbf{in-degree} indicates an information ``gatherer,'' a component that integrates inputs from many different predecessors to perform its computation.

\paragraph{Centrality measures}
To understand the efficiency of information flow and identify critical pathways, I compute centrality measures.
The \textbf{betweenness} centrality of a node $v$ measures how often it lies on the shortest path between other pairs of nodes: $C_B(v) = \sum_{s \neq v \neq t} \frac{\sigma_{st}(v)}{\sigma_{st}}$, where $\sigma_{st}$ is the total number of shortest paths from $s$ to $t$. Nodes with high betweenness centrality act as crucial ``gatekeepers'' or ``bridges'' that control information flow, and their emergence may signify the formation of efficient computational circuits.
The \textbf{closeness} centrality of a node $v$ measures its average farness (inverse distance) to all other nodes: $C_C(v) = \frac{|V|-1}{\sum_{u \neq v} d(v, u)}$, where $d(v, u)$ is the shortest path distance. Nodes with high closeness centrality are ``spreaders'' that can propagate information efficiently throughout the network. An increase in average closeness centrality would suggest that the network is becoming more integrated and efficient at global communication.

\section{Experimental Setup}
\label{sec:exp-setup}

\subsection{Model}
I utilize the \texttt{pythia-14m} model from the EleutherAI suite~\cite{biderman2023pythia}, a 6-layer Transformer with 4 attention heads per layer (1.2M parameters). This model was chosen for two primary reasons: its open-source nature and, most importantly, the public availability of 143 training checkpoints. This extensive set of checkpoints allows for a fine-grained, longitudinal analysis of the network's structural evolution. I interface with the model and its internal components using the \texttt{transformer-lens} library~\cite{nanda2022transformerlens}.

\subsection{Induction task design}
To investigate the formation of circuits related to in-context learning, I focus on a canonical \textbf{induction task}. I provide the model with a fixed input text that contains repeated name patterns. The model's objective is to complete the final, truncated name based on the patterns established earlier in the context. Successful prediction requires the model to identify prior occurrences of the pattern and apply that knowledge, a hallmark of induction.

The specific input text is:
\begin{quote}
\small
    Mr. and Mrs. Dursley, of number four, Privet Drive, were proud to say
    that they were perfectly normal, thank you very much. They were the last
    people you'd expect to be involved in anything strange or mysterious,
    because they just didn't hold with such nonsense.
    Mr. Dursley was the director of a firm called Grunnings, which made
    drills. He was a big, beefy man with hardly any neck, although he did
    have a very large mustache. Mrs. Dursley was thin and blonde and had
    nearly twice the usual amount of neck, which came in very useful as she
    spent so much of her time craning over garden fences, spying on the
    neighbors. The Durs
\end{quote}
The model's task is to predict the next token, which should be `ley'.

\subsection{Graph Construction and Analysis Protocol}
The analysis follows a systematic protocol for each available training checkpoint. Based on the computed causal influence scores, I construct a set of directed, weighted graphs, each corresponding to a different similarity threshold $\tau \in \{0.1, 0.3, 0.5, 0.7, 0.9, 1.0\}$. A lower threshold results in a sparser graph capturing only the strongest interactions, while $\tau=1.0$ represents a fully connected graph of all potential forward-passing influences. For each generated graph, I then compute the suite of complex network metrics outlined in Section~\ref{sec:methodology} using the \texttt{NetworkX} library.

This protocol yields a time-series dataset where each training checkpoint is associated with a set of graph-theoretic metrics and a corresponding performance score. The subsequent analysis aims to investigate whether the evolution of these metrics can reveal phase changes and correlate structural shifts in the component-graph with the model's learning dynamics.

\section{Results}
\label{sec:results}
In this section, I analyze the training dynamics by examining the evolution of key graph metrics. To link these structural changes to learning, I overlay the logit of the correct token on each plot, which serves as a proxy for the model's performance on the induction task.

\begin{figure*}[h!]
    \centering
    \begin{subfigure}{0.32\textwidth}
        \centering
        \includegraphics[width=\linewidth]{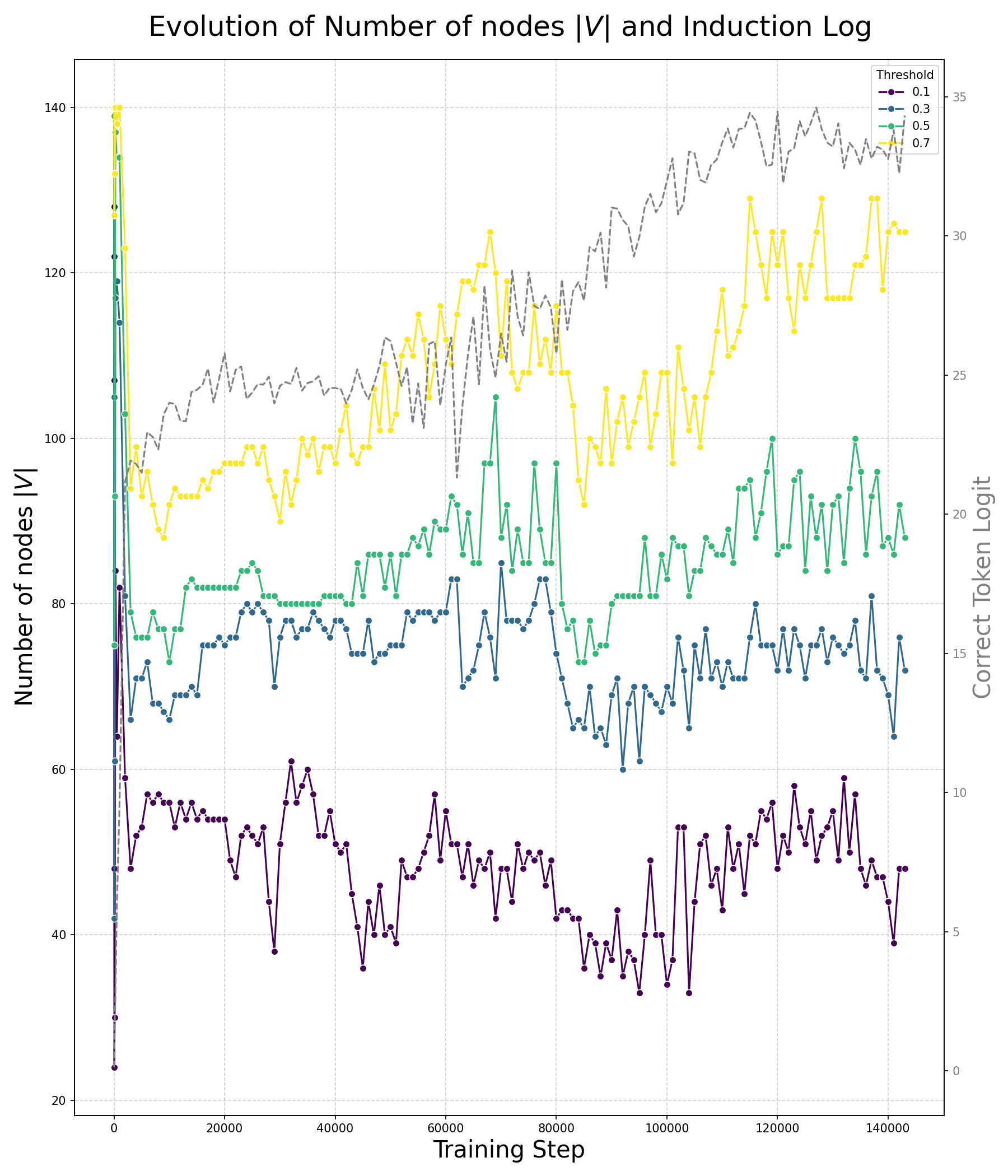}
        \caption{Evolution of the number of nodes.}
        \label{fig:nodes_evo}
    \end{subfigure}
    \begin{subfigure}{0.32\textwidth}
        \centering
        \includegraphics[width=\linewidth]{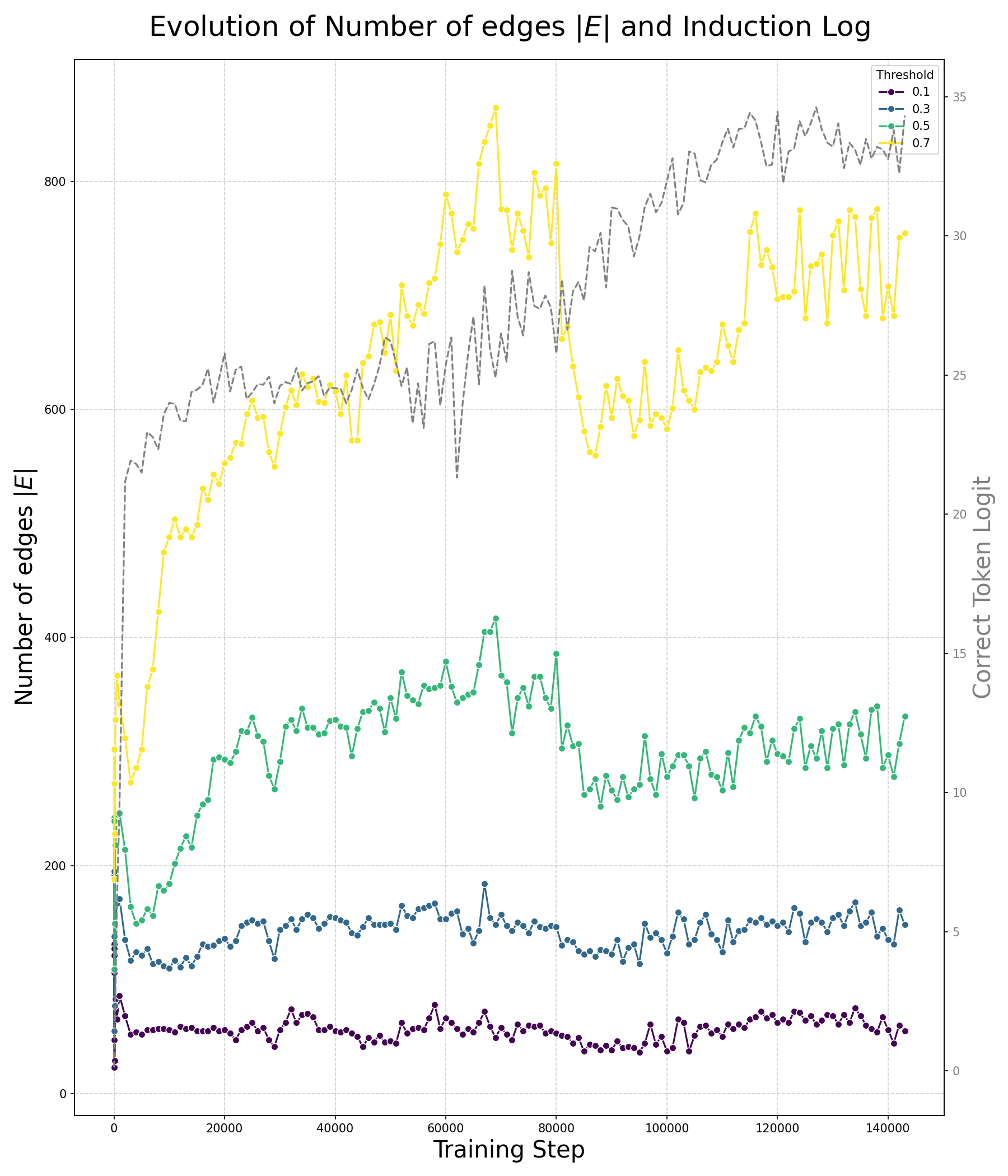}
        \caption{Evolution of the number of edges.}
        \label{fig:edges_evo}
    \end{subfigure}
    \begin{subfigure}{0.32\textwidth}
        \centering
        \includegraphics[width=\linewidth]{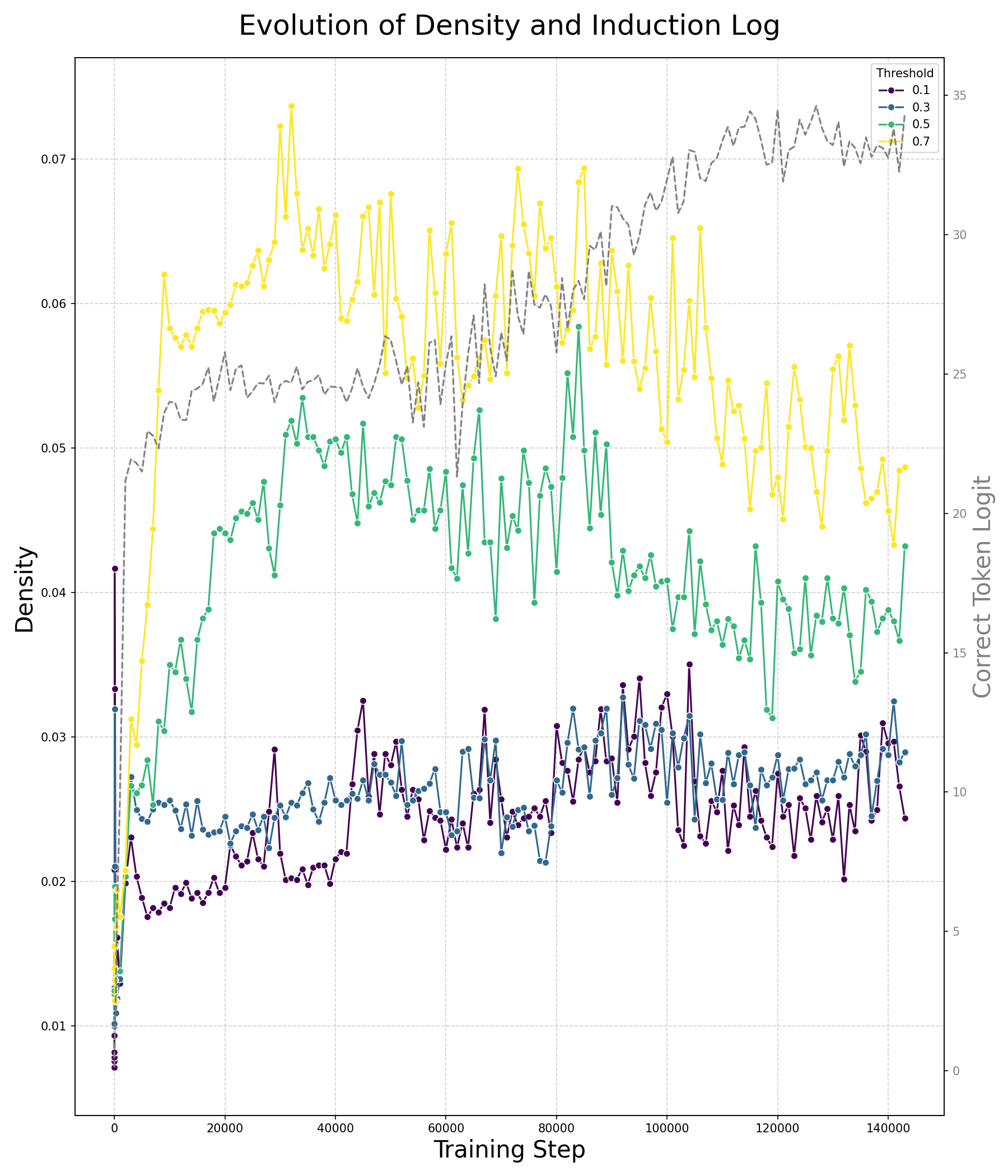}
        \caption{Evolution of network density.}
        \label{fig:density_evo}
    \end{subfigure}
    
    \vspace{5mm} 
    
    \begin{subfigure}{\textwidth}
        \centering
        \includegraphics[width=\textwidth]{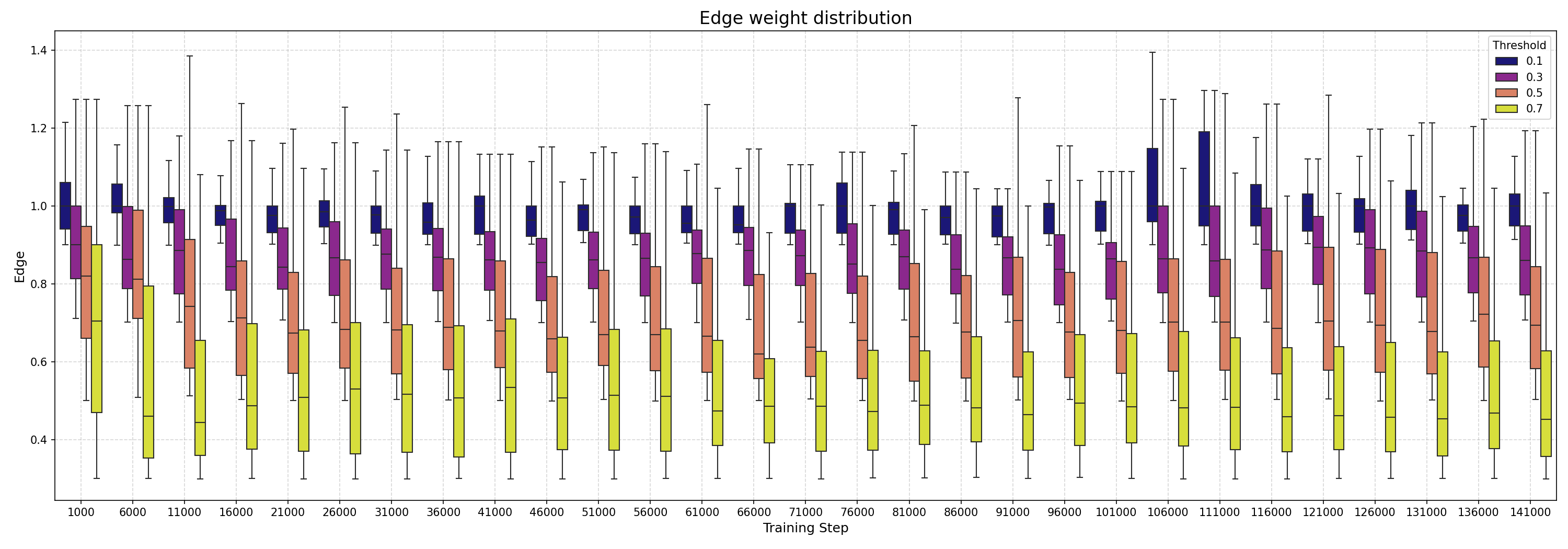}
        \caption{Evolution of edge weight distribution.}
        \label{fig:weight_evo}
    \end{subfigure}
    
    \caption{Evolution of topological properties over training for different thresholds, with the correct token logit (gray, right y-axis) overlaid as a performance proxy.}
    \label{fig:topo_evolution}
\end{figure*}

\paragraph{$|V|, |E|$ and density}
Analysis of the graph's basic topological properties reveals a dynamic evolution that occurs in two distinct phases, mirroring the model's learning trajectory as indicated by the correct token logit.
The initial training phase (0 to \textasciitilde20k steps) is characterized by a rapid, explosive growth in both the number of active nodes ($|V|$) and edges ($|E|$), particularly for higher thresholds (e.g.$\tau=0.7$, see Figure~\ref{fig:topo_evolution}a,b,c). This initial ``burst'' directly correlates with the steep increase in the correct token logit, suggesting an exploratory period where the model establishes a wide array of potential communication pathways to solve the task. This is immediately followed by a consolidation phase, where the number of active nodes begins to decrease, indicating a pruning of less effective components and a refinement of the network toward greater computational efficiency.

A second, more subtle phase of structural change occurs around the 60k-80k step mark. Here, another period of network pruning is visible as both $|V|$ and $|E|$ trend downward. Crucially, this event coincides with a shift in the learning dynamics: the correct token logit exhibits a new period of gradual improvement after this structural refinement. This suggests that the network is not learning the task from scratch, but is instead discovering and solidifying more specialized and efficient circuits by discarding redundant pathways. While the network density fluctuates without a clear trend, the distinct evolution of node and edge counts provides a structural signature of the model's learning process: from initial, broad exploration to subsequent phases of targeted refinement and circuit specialization.

\paragraph{Edge weight distribution}
Consistent with findings in the literature, the edge weight distribution becomes increasingly long-tailed over the course of training. This reflects the network's specialization, where a majority of inter-component connections remain weak, while a few critical pathways emerge with very high weights. This suggests the formation of efficient circuits built around a small number of strong ``information highways'' (see Figure ~\ref{fig:topo_evolution}d).

For the remainder of this results section, I present both distribution visualizations and grids highlighting the components with the highest values for each specified metric. Components are selected if their values exceed the 95th percentile of the distribution at the given timestep. The grid plots are shown only for $\tau = 0.7$, which represents a ``sweet spot'' where a sufficient number of connections exert meaningful influence on one another. This choice is motivated by how the graphs have been built, making metrics distributions highly skewed with many outliers in the upper tail.

\begin{figure*}[h!]
    \centering
    \includegraphics[width=\textwidth]{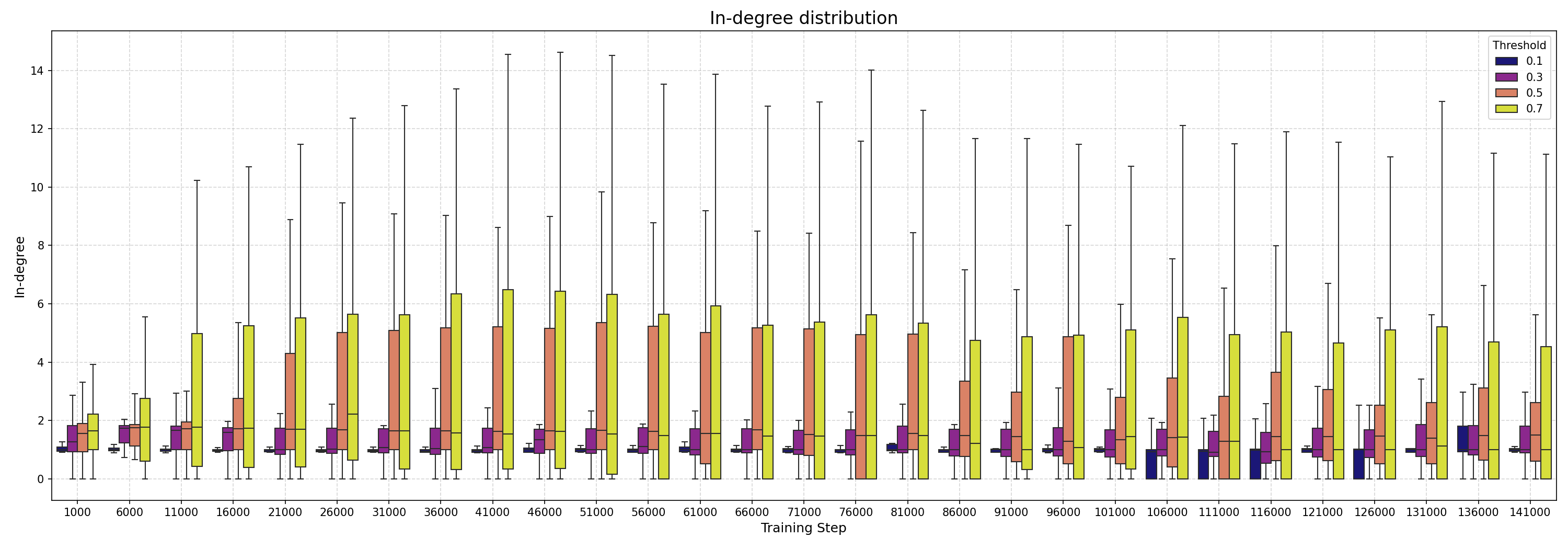}
    \includegraphics[width=\textwidth]{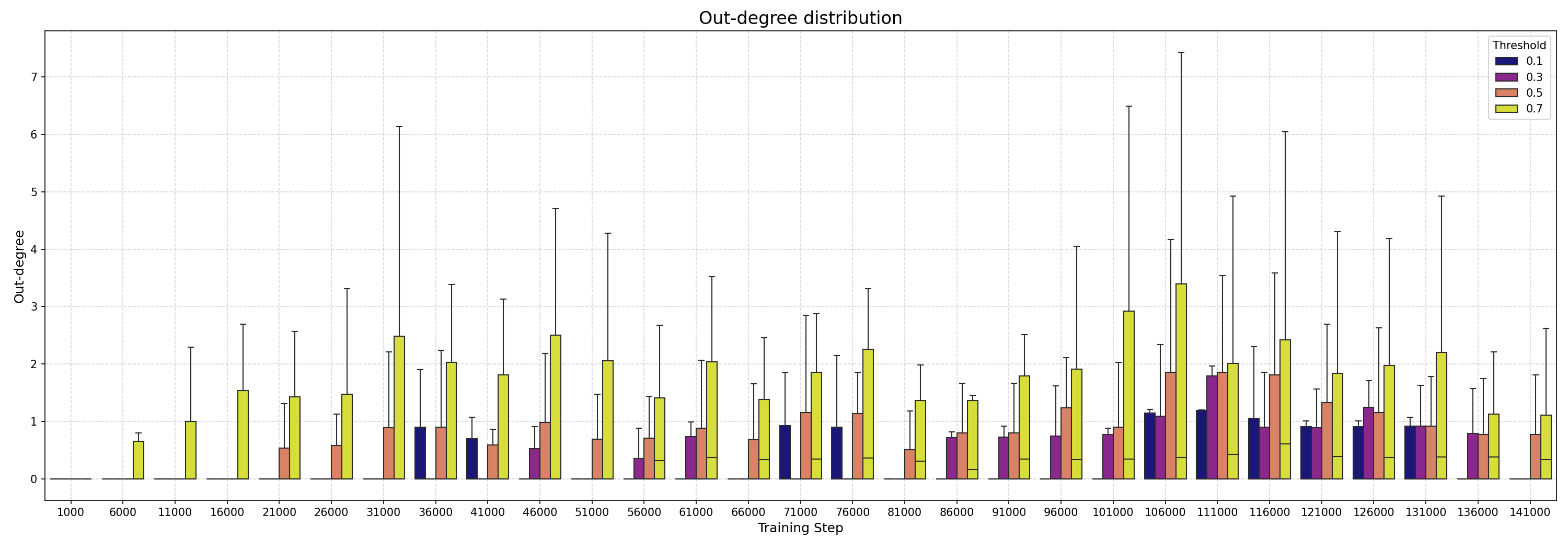}
    \caption{Boxplots showing the evolution of the In-degree (top) and Out-degree (bottom) distributions over all training steps, grouped by threshold.}
    \label{fig:degree_dists}
\end{figure*}

\paragraph{In-degree and information gatherers}
The in-degree distribution (Figure \ref{fig:degree_dists}a) reveals a clear trend towards specialization. It evolves from a relatively symmetric distribution at the start of training to one with a pronounced right-skewed tail. This indicates the emergence of ``information gatherers''---a minority of components that learn to integrate information from a large and diverse set of preceding components in the residual stream. 

\begin{figure*}[h!]
    \centering
    \includegraphics[width=\textwidth]{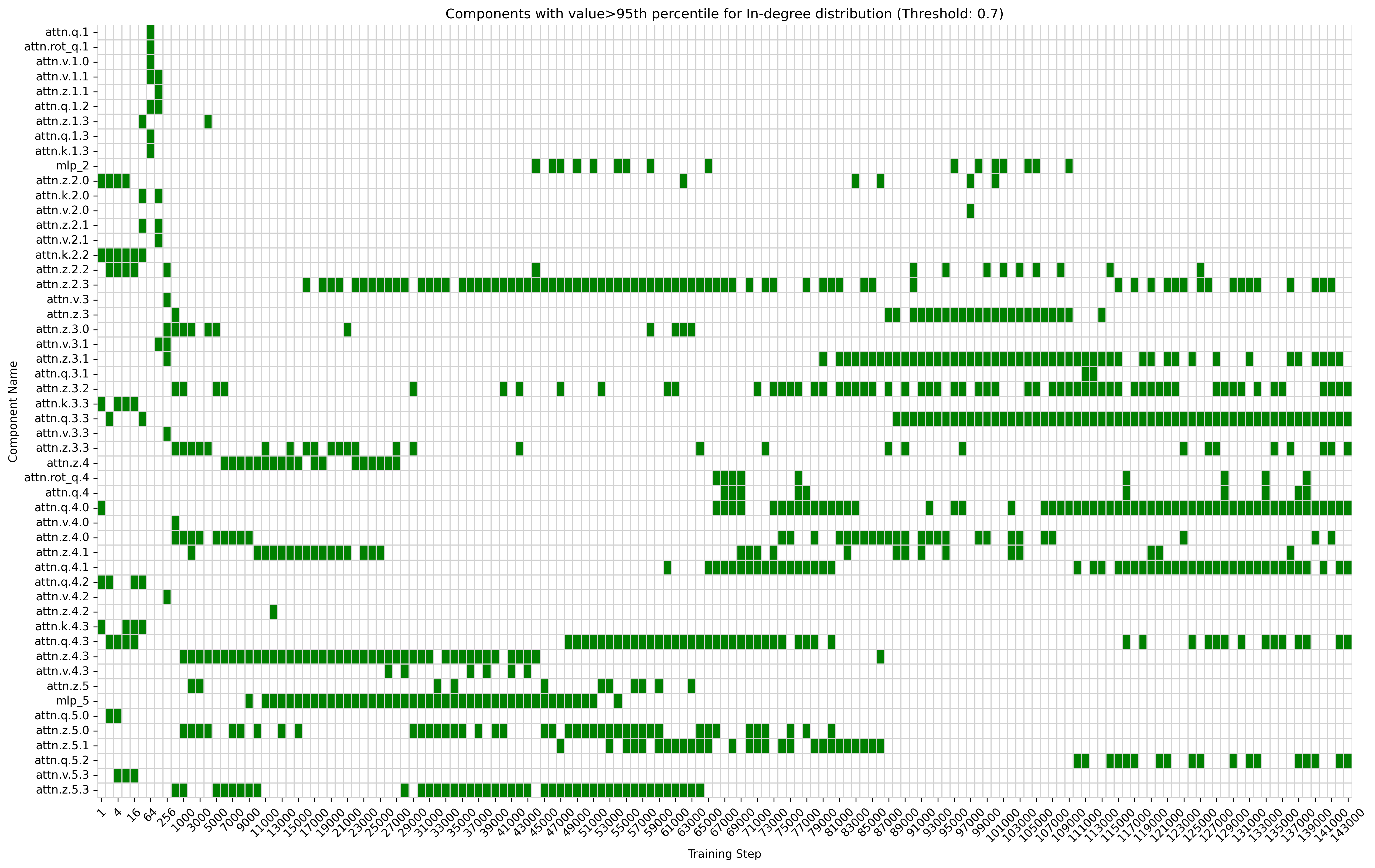}
    \includegraphics[width=\textwidth]{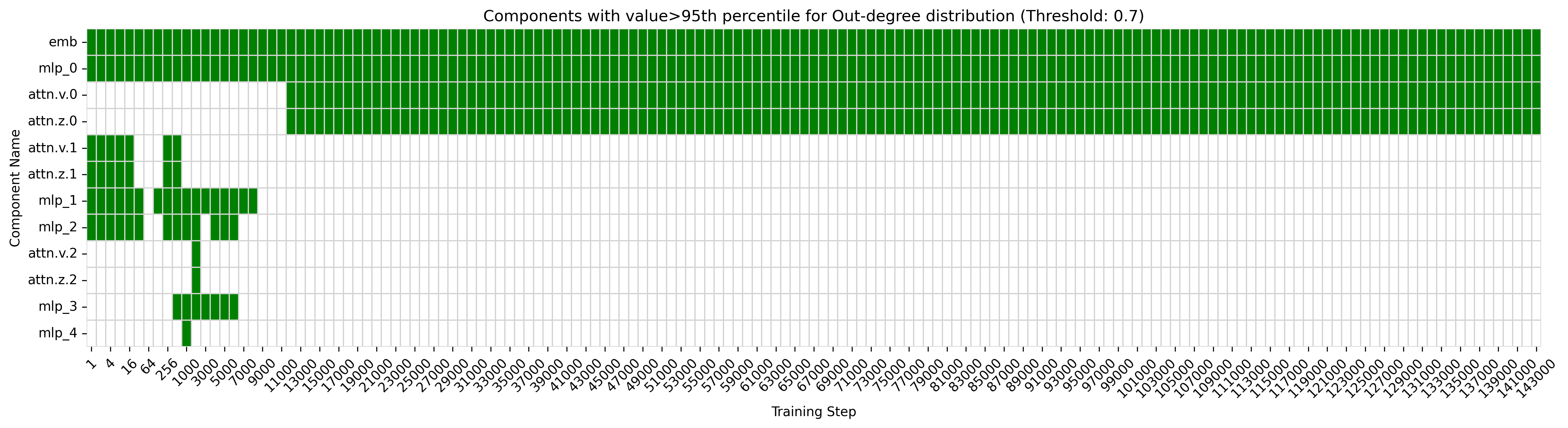}
    \caption{Heatmaps identifying components with In-degree (top) and Out-degree (bottom) values in the >95th percentile at each training step for $\tau=0.7$. Green indicates that a component is a high-degree hub or authority at that step.}
    \label{fig:degree_percentiles}
\end{figure*}

Crucially, the heatmap in Figure \ref{fig:degree_percentiles}a shows that the set of high-in-degree authorities is not static; instead, it \textit{reconfigures at key junctures that align with the learning phases}. During the initial consolidation period (up to to \textasciitilde1 steps), attention heads are the first information gathering units to stand out. Then, from step \textasciitilde10k to \textasciitilde80k other components seem to read from the residual stream, for example \texttt{attn.z.2.3}, attention heads from the 4th and 5th layer, and even MLP modules like \texttt{mlp\_2} and \texttt{mlp\_5}. However, a third reconfiguration occurs after the \textasciitilde80k-step mark, where some previously dominant components cede prominence and a new set of authorities, such as heads in layer 3, emerges. This dynamic reallocation suggests that as the model refines its solution, it is not merely strengthening a fixed circuit but is actively discovering more efficient computational pathways.

\paragraph{Out-degree and information spreaders}
The out-degree analysis reveals a strikingly stable hierarchy of ``information spreaders''. The heatmap in Figure \ref{fig:degree_percentiles}b shows that a small, consistent set of components act as the primary hubs for almost the entire training process after the first \textasciitilde10k steps: \texttt{emb} and \texttt{mlp\_0}. Other units are sharing information via the residual stream in the initial training phase, but then are replace by two main units (\texttt{attn.v.0} and \texttt{attn.z.0}). This suggests the model quickly converges on a fundamental information flow pattern: foundational features are computed early and broadcast widely through the residual stream. Later-layer components, which showed high in-degree, act more as specialized processors that synthesize this broadcasted information rather than as primary broadcasters themselves. The out-degree distribution (Figure \ref{fig:degree_dists}b) supports this, showing that while some outliers emerge, the bulk of components maintain a low out-degree throughout training. One last note from Figure \ref{fig:degree_dists}b is the emergence of some weaker information spreader showing up from the \textasciitilde80k step.

\begin{figure*}[h!]
    \centering
    \includegraphics[width=\textwidth]{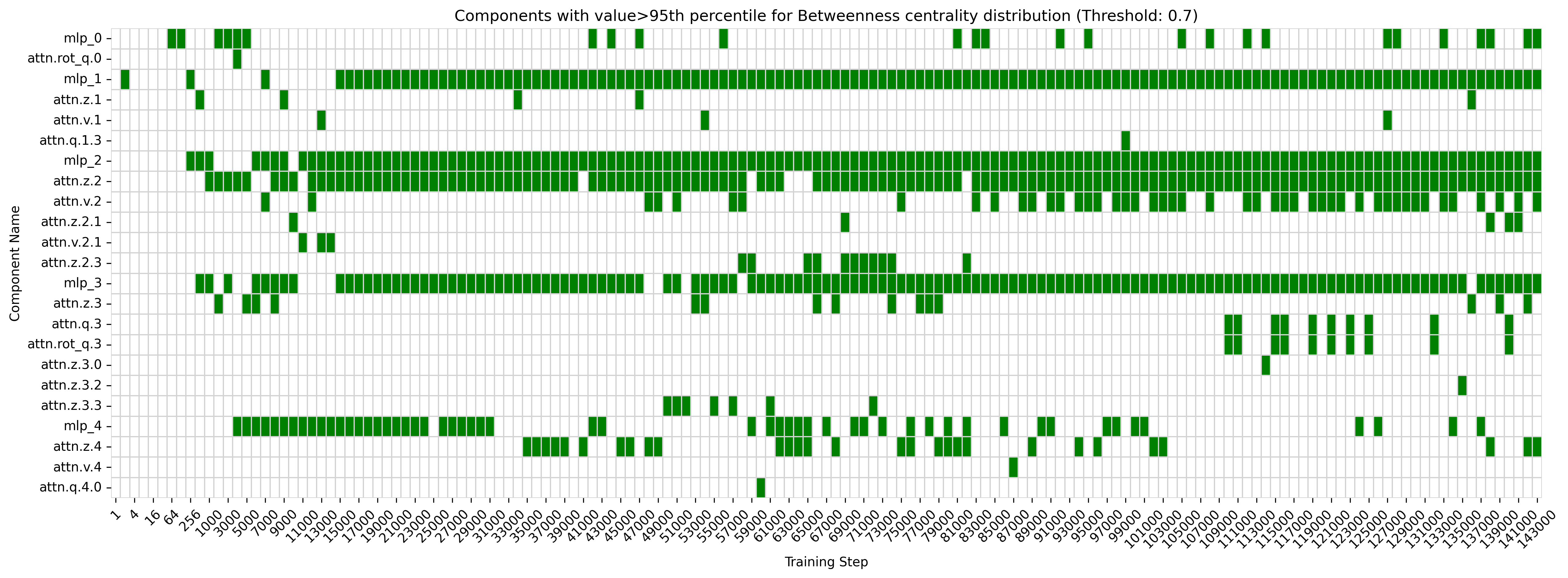}
    \caption{Heatmap identifying components with betweenness centrality values in the >95th percentile at each training step for $\tau=0.7$. Green indicates that a component is a critical information gatekeeper at that step.}
    \label{fig:betweenness_heatmap}
\end{figure*}

\paragraph{Betweenness centrality and information gatekeepers}
Betweenness centrality identifies components that act as critical ``gatekeepers'' or bridges, controlling information flow by lying on many of the shortest computational pathways. In my analysis, the vast majority of components exhibit a betweenness centrality of zero, indicating that only a small, specialized subset of nodes form these crucial bridges. Therefore, I focus the analysis on the heatmap in Figure~\ref{fig:betweenness_heatmap}, which tracks the components with centrality values in the >95th percentile.

The results reveal a stable core of information gatekeepers that emerges early (around step 1) and persists throughout training. These include the MLP blocks from the middle layers (\texttt{mlp\_1}, \texttt{mlp\_2}, \texttt{mlp\_3}) and the output of the attention heads in layer 2 (\texttt{attn.z.2}). More interestingly, the heatmap also shows a dynamic rewiring of these pathways. Before the ~100k step mark, components like \texttt{mlp\_4} and the outputs of attention heads in layers 3 and 4 (\texttt{attn.z.3}, \texttt{attn.z.4}) are frequently central. However, after this point, they are largely superseded by a new set of gatekeepers, including several attention components in layer 3.

This dynamic behavior provides strong evidence that the network is not merely strengthening fixed pathways but actively re-routing its information flow. The shift in which components act as gatekeepers suggests the model is discovering new and more efficient computational circuits as it refines its ability to perform the induction task.

\begin{figure*}[h!]
    \centering
    \includegraphics[width=\textwidth]{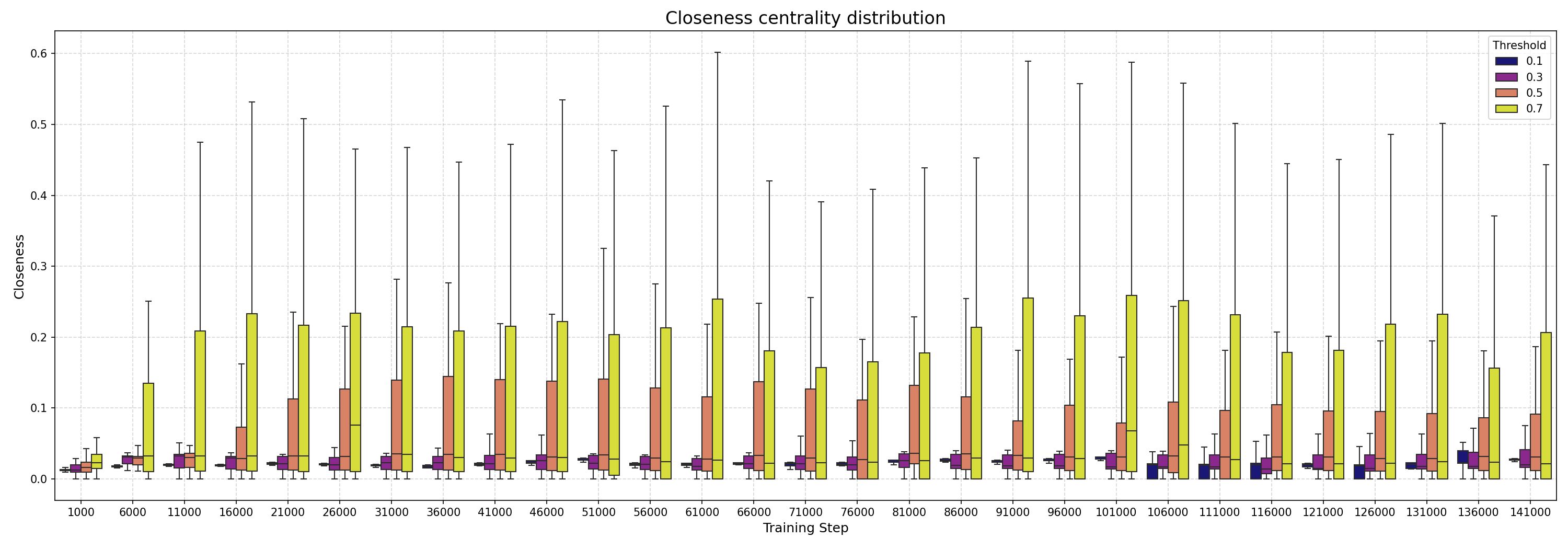}
    \caption{Boxplot showing the evolution of the Closeness centrality distribution over all training steps, grouped by threshold.}
    \label{fig:closeness_dist}
\end{figure*}

\paragraph{Closeness centrality and information spreading efficiency}
Closeness centrality identifies components that act as efficient information spreaders, measuring how quickly they can propagate information to all other reachable nodes in the network. The distribution plots in Figure~\ref{fig:closeness_dist} reveal a clear and significant trend: more and more nodes acquire higher closeness centrality for all thresholds throughout training. This indicates that the component-graph as a whole becomes progressively more integrated and globally efficient, reducing the average path length between any two components and allowing for faster communication.

\begin{figure*}[h!]
    \centering
    \includegraphics[width=\textwidth]{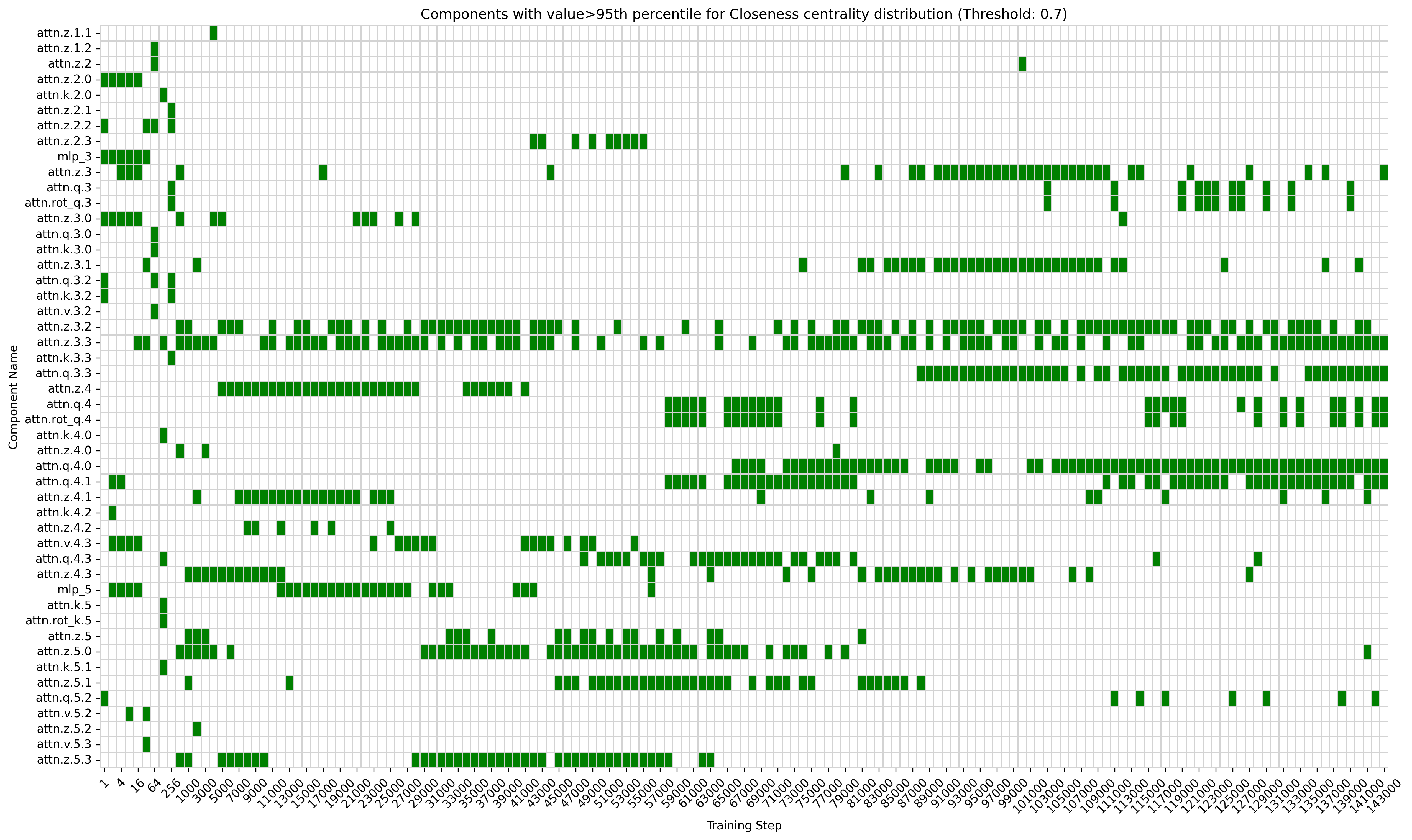}
    \caption{Heatmap identifying components with closeness centrality values in the >95th percentile at each training step for $\tau=0.7$. Green indicates that a component is an efficient information spreader at that step.}
    \label{fig:closeness_heatmap}
\end{figure*}

The heatmap in Figure~\ref{fig:closeness_heatmap} provides a granular view of this process, revealing a dynamic evolution that can be segmented into three distinct phases. The first phase (0 - 10k steps) is characterized by a transient set of components, primarily attention heads in the middle layers, briefly emerge as central hubs. These components quickly fade in importance, suggesting they represent an initial, exploratory attempt by the model to establish global communication pathways. During a second phase (10k - 80k steps), a new circuit is formed. Components in the later layers, particularly \texttt{mlp\_5} and several heads in layer 5, become the dominant information spreaders. Eventually (80k steps onwards), a further optimization occurs. \texttt{mlp\_3} and attention heads from the 4th layer emerge as a stable central hub. The only units which consistently remain central are \texttt{attn.z.3.2} and \texttt{attn.z.3.3}.

\section{Conclusion}
\label{sec:conclusion}

In this project, I investigated the feasibility of characterizing LLM training dynamics by representing the model as an evolving complex network of its computational components. By constructing a time-series of graphs based on causal influence and analyzing their topological properties, I sought to determine if this macroscopic view could reveal the structural signatures of learning, particularly the phase change associated with the formation of induction heads.

The results support this hypothesis. The analysis revealed that the component-graph undergoes a clear, multi-phase evolution. This process begins with an exploratory phase of rapid, widespread connectivity, followed by a consolidation phase where the network prunes inefficient components. Finally, a prolonged refinement phase sees the network discovering new, more efficient information pathways, evidenced by the dynamic reconfiguration of key information gatherer (high in-degree) and gatekeeper (high betweenness centrality) components. In contrast, a stable hierarchy of information spreader (high out-degree) hubs, primarily the embedding and early-layer MLPs, is established early and persists, suggesting a fundamental pattern of information broadcasting and subsequent specialized processing.

Taken together, these findings demonstrate that a component-level graph representation provides a powerful and interpretable lens for understanding how a Transformer's internal architecture self-organizes. The evolution of metrics like degree and centrality offers a tangible, macroscopic signature of the model's learning process, moving from broad exploration to the formation and refinement of specialized computational circuits.

\subsection{Limitations and Future Work}
This study provides a proof-of-concept, and several limitations should be acknowledged, which also point to promising directions for future research.
\paragraph{Model and scale} The analysis was conducted on a single, small model (\texttt{pythia-14m}). Future work should apply this methodology to larger models to investigate how these network dynamics change with scale.
\paragraph{Task specificity} The component-graph was generated in the context of a single, canonical induction task. It remains an open question whether different tasks (e.g., factual recall, translation) would induce the formation of structurally distinct networks. A comparative analysis across multiple tasks could reveal how models develop task-specific circuits.
\paragraph{Input dependency} The graph was constructed using a single input sentence. While chosen to be representative, the graph structure is likely dependent on the input tokens. Future work could explore the robustness of these structures by averaging influence scores over a dataset of inputs.
\paragraph{Ablation method} My choice of zero-ablation is one of several possible intervention strategies. Other techniques, such as mean-ablation or resampling from other inputs, could yield different insights into component influence and are an important avenue for further investigation.

Ultimately, this work lays the groundwork for using CNT as a scalable tool in mechanistic interpretability. Future research could use pathfinding algorithms on these graphs to automatically identify candidate circuits for more detailed, micro-level analysis, bridging the gap between macroscopic dynamics and the specific mechanisms that give rise to them.

\bibliography{main}

\end{document}